%% file: naaclhlt2018.tex
\documentclass[11pt,a4paper]{article}
\usepackage[hyperref]{naaclhlt2018}
\usepackage{times}
\usepackage{latexsym}
\usepackage{url}

\usepackage{booktabs} % For formal tables

\usepackage{textcomp}
\usepackage{latexsym}
\usepackage{graphicx} % "demo" option just for this example
\usepackage{microtype}
\usepackage{pifont}
\usepackage[font=small]{caption}
\usepackage[]{subcaption}
\usepackage{verbatim}
\usepackage{framed}
\usepackage{color, colortbl}
\usepackage[colorinlistoftodos]{todonotes}
\usepackage{amssymb}
\usepackage{amsmath}
\usepackage{breqn}
\usepackage{capt-of}
\usepackage{multirow}
\usepackage{algorithmicx}
\usepackage{algorithm}
\usepackage{algcompatible}
\usepackage{paralist}
\usepackage{enumitem}
\usepackage{balance}
\usepackage{tabularx}
\usepackage[all]{nowidow}

\definecolor{high}{HTML}{FFCE8E}
\aclfinalcopy % Uncomment this line for the final submission
\def\aclpaperid{391} %  Enter the acl Paper ID here

\title{Simple Models for Word Formation in English Slang}

\author{Vivek Kulkarni \\
    Department of Computer Science \\ 
  	University of California, Santa Barbara \\
  {\tt vvkulkarni@cs.ucsb.edu} \\\And
  William Yang Wang \\
    Department of Computer Science \\ 
  	University of California, Santa Barbara \\
  {\tt william@cs.ucsb.edu} \\}

\date{}

\begin{document}
\maketitle
\begin{abstract}
We propose generative models for three types of extra-grammatical word formation phenomena abounding in English slang: \emph{Blends, Clippings, and Reduplicatives}. Adopting a data-driven approach coupled with linguistic knowledge, we propose simple models with state of the art performance on human annotated gold standard datasets. Overall, our models reveal insights into the generative processes of word formation in slang -- insights which are increasingly relevant in the context of the rising prevalence of slang and non-standard varieties on the Internet. 
\end{abstract}
\input{introduction}
\input{dataset}
\input{methods}
\input{experiments}
\input{relatedwork}
\input{conclusion}
\input{acknowledgements}
%\bibliography{naaclhlt2018}
\bibliography{naaclhlt2018}
\bibliographystyle{acl_natbib}

%\appendix

%\section{Supplemental Material}
%\label{sec:supplemental}

%\section{Multiple Appendices}
%\dots can be gotten by using more than one section. We hope you won't
%need that.

\end{document}

%% file: introduction.tex
\section{Introduction}
Linguistic analysis of slang has traditionally received little attention with some arguing that research on slang be assigned to an \emph{``extra-linguistic darkness''} \cite{labov1972some}. However, \citet{eble2012slang} argues that the emergence of social media has predominantly increased the usage of slang and non standard forms and \emph{``slang is now worldwide the vocabulary of choice of young people''}\footnote{While the definition of slang is a controversial issue, we adopt a broad definition including non-standard expressions.}. This increasing pervasiveness has recently motivated research on slang and the linguistic phenomena it manifests.
Most notable are the works of \cite{mattiello2005pervasiveness,mattiello2008introduction,mattiello2013extra}, who argues that slang exhibits extra-grammatical properties that distinguish it from the standard form. 

Specifically, linguistic phenomena like \emph{alphabetisms}, \emph{blending}, \emph{clippings}, and \emph{reduplicatives} abound in slang (see Table \ref{tab:crown_jewel})\footnote{While such phenomena are likely present in several languages, in this work we restrict ourselves to slang in English.}.
Note the rich and varied word formation patterns ranging from simple abbreviations like \texttt{dink} to more complex combinations like \texttt{lambortini}, a blend of \texttt{lamborghini} and \texttt{martini}. 
Note further, that even within a particular class like \textsc{Blends} there are variations in what portions of the components are retained. 
\begin{table}[t!]
\small
\centering
\begin{tabularx}{\columnwidth}{l|l|l}
\textbf{Word} & \textbf{Derived From} & \textbf{Type} \\ \midrule
dink & double income no kids & alphabetism \\  
lambortini & lamborghini + martini & blend \\
diamat & dialectical + materialism & blend \\
tude & attitude & clipping (fore) \\ 
brill & brilliant & clipping (back) \\ 
teenie-weenie & teenie & reduplicative \\
yik-yak & yik & reduplicative \\ \midrule
\end{tabularx}
\caption{Sample words depicting word formation in slang. Note the rich variation across different types.}
\label{tab:crown_jewel}
\end{table}
These word formation mechanisms are not only attractive from a linguistic standpoint in deepening our understanding of slang but also have applications spanning the development of rich conversational agents and tools like brand name generators \cite{ozbal2012computational}. While such phenomena have been qualitatively studied by \citet{mattiello2008introduction,mattiello2013extra}, computational models for their generation have not been proposed. 

In this paper, we propose the first simple models for generating blends, clippings, and reduplicatives\footnote{While there has been some prior work on computationally modeling blends, generative models for clippings and reduplicatives have not been outlined.}. 
Our models incorporate linguistic insights coupled with data-driven analysis to model the above phenomena. In line with ``Occam's razor'', we strive for simplicity. 
The simplicity of our models not only implies better generalization, more robust estimation of parameters in the wake of small dataset sizes, and better interpret-ability but also yields state of the art performance. 

Specifically, we show that by exploiting structural constraints, blend formation can be modeled as a simple \emph{sequence labeling} problem 
as opposed to prior work which models it as a \emph{general sequence to sequence} problem. This view enables the use of a simple LSTM model to yield competitive performance. Similarly, we propose the first probabilistic generative models for clippings and reduplicatives effectively incorporating phonetic constraints. In a nutshell, our contributions are:
\begin{enumerate}
    \item \textbf{Generative models.}  We propose simple models for generating blends, clippings and, reduplicatives with state of the art performance.
    \item \textbf{Linguistic Insights.} We reveal linguistic insights into these phenomena which we incorporate into the generative models.
    \item \textbf{Resources.} We release all our models and the compiled datasets to aid further research \footnote{Code and data with reproducible results is available at \texttt{https://github.com/viveksck/simplicity}}. 
\end{enumerate}

%% file: dataset.tex
\section{Datasets and Definitions}
Here, we define the extra-grammatical morphological phenomena modeled and describe the datasets used for our experiments and analysis.
\citet{mattiello2013extra} argues that slang exhibits extra-grammatical morphological properties that distinguish them from the standard variety and identified four broad word formation phenomena\footnote{These categories are not exhaustive and several slang words and non-standard expressions do not fall into any of the above categories.} described below:
\begin{enumerate}
	\item \textbf{Alphabetisms} are shortenings of a multi-word sequence. 
	Examples include \texttt{lol} from \texttt{laugh out loud} or \texttt{YOLO} from \texttt{you only live once}. 
	They can be further sub-categorized into two types based on their pronunciation although the distinction may not always be clear: (a) \emph{Acronyms} are pronounced using the regular reading rules (for example. YOLO) (b) \emph{Initialisms} are pronounced letter by letter (for example. BBC). 
	\item \textbf{Blends} or portmanteaus, are formed by merging parts of existing words.
	For example, \texttt{edutainment} is a blend of \texttt{education} and \texttt{entertainment}.  Prior work notes that blend formation does not exhibit rigid rules but only demonstrates affinities towards certain patterns of formation \cite{mattiello2013extra} suggesting learning based approaches to modeling blends \cite{deri2015make,gangal2017charmanteau}. 
	\item \textbf{Clippings} are constructed by  shortening words (lexemes). For example, \texttt{berg} is a clipping of \texttt{iceberg}, \texttt{gym} is a clipping of \texttt{gymnasium} and \texttt{ammo} is a clipping of \texttt{ammunition}. 
	Based on the portion that is being clipped, clippings are sub-categorized into three types: (a) \textsc{Back} clipping where the beginning of the word (lexeme) is retained (like \texttt{brill} from \texttt{brilliant}) (b) a \textsc{Fore} clipping, where the end of a word is retained (like \texttt{choke} from \texttt{artichoke}) and (c) A \textsc{Compound} clipping (\texttt{adman}), a clipping of a compound word (\texttt{advertisment man}). 
	%Most clippings are usually back-clippings while forward clippings are relatively rare.
	\item \textbf{Reduplicatives} are word pairs constructed by either repeating a word (\texttt{boo boo}) or by alternating certain vowels or consonants so that they are phonologically similar (\texttt{clickety-clackety, teenie-weenie, itsy-bitsy}). 
\end{enumerate}

%Although the above classes have been qualitatively analyzed \cite{mattiello2008introduction,mattiello2013extra}, explicit generative models capturing these linguistic phenomena have not been proposed. 
In our work, we propose generative models using a data-driven approach towards generating blends, clippings ,and reduplicatives.
We do not consider generative models for alphabetisms since a majority of them can be trivially generated by picking the first letter of each word making up the acronym (for example. \texttt{laugh out loud} $\rightarrow$ \texttt{lol}).
We now outline the datasets considered:
\begin{enumerate}
    \item \textbf{Blends.} We consider a gold standard dataset $\mathcal{D}_{knight}$ of $400$ blends constructed by \cite{deri2015make} from Wikipedia as well as a larger list of $1624$ blends manually compiled by \cite{gangal2017charmanteau} called $\mathcal{D}_{large}$, a superset of $\mathcal{D}_{knight}$. We define $\mathcal{D}_{blind}=\mathcal{D}_{large}-\mathcal{D}_{knight}$. 
    \item \textbf{Clippings.} We consider a list of $576$ human curated clippings constructed by \citet{mattiello2013extra} for our analysis of clippings. These were manually collected from a variety of sources including prior work and dictionaries like the \emph{Oxford English Dictionary} and the \emph{Merriam-Webster online dictionary}.  
    \item \textbf{Reduplicatives.} We consider a dataset of a total of $337$ reduplicatives manually constructed by \citet{mattiello2013extra} also collected from prior work and online dictionaries\footnote{We note that the dataset released by \cite{mattiello2013extra} is the largest human curated dataset of slang word formation we are aware of. Furthermore, large crowd sourced data-sets like Urban Dictionary are a poor fit for such modeling since they do not explicitly contain annotations by linguistic experts.}.  
\end{enumerate}
%All of the above datasets will be made publicly available to aid further research into modeling these linguistic phenomena.

%% file: methods.tex
\section{Models and Methods}
We now describe our proposed models to generate blends, clippings, and reduplicatives.  We precisely formulate the problem,  specify our models and comprehensively outline our evaluation.
\subsection{Blends}
\paragraph{Problem Formulation}
Given a string $C=C_{1}\#C_{2}$, consisting of components $C_{1}$ and $C_{2}$, we seek to combine them to yield the blend $B$. 
For example, given $C_1=\texttt{brad}$ and $C_2=\texttt{angelina}$ such that $C=\texttt{brad\#angelina}$, we seek to generate the blend \texttt{brangelina}. 

\paragraph{Existing Models}  \cite{deri2015make} proposed a model to generate blends using multi-tape finite state transducers.  Most recently, \cite{gangal2017charmanteau} (the current state of the art) model this as a general sequence to sequence learning problem and propose a neural encoder-decoder architecture with attention to outperform the model by \cite{deri2015make}.  However, this model fails to effectively exploit the inherent structure and linguistic constraints of blending. One implication is an exploration of an overly complex hypothesis space with a small amount of training data making it harder to generalize. 
A second implication is that the decoding phase uses exhaustive generation. In fact, the best model \emph{exhaustively} generates all candidate strings, where the first part is a prefix of $C_{1}$ and the second part is a suffix of $C_{2}$ and scores them to pick the best candidate while using a backward model learned to generate the components given the blend.  In contrast, we propose a more straightforward model that explicitly incorporates inherent linguistic constraints entirely obviating the need for decoding using exhaustive candidate generation yet yielding competitive performance. 

\subsubsection{Linguistic Insights into Blend Formation}
\label{sec:insights}
While one can model the problem of learning to blend as a variable length sequence to sequence learning problem (akin to machine translation), we argue that incorporating structural constraints yields a different view of modeling the problem that can enable better generalization given the small amount of training data. We motivate this by observing the following constraints:
\begin{enumerate}
    \item \textbf{Blend length and vocabulary constraints.} First, we observe that a majority of blends ($99.0\%$ in $D_{knight}$ and $92.4\%$ in $D_{blind}$) are formed by using only characters present in the original components.  Second, the length of the blends in these cases is at-most the length of the components.
    \item \textbf{Fixed length input output representation.} The blend $B$ can thus be encoded as a string of the exact same length as $C$ by noting that $B$ only contains characters copied from $C$ or deleted from $C$. Specifically, we represent $B$ by $E(B)$ denoting the sequence of \emph{copy} (\textbf{C}) and \emph{delete} (\textbf{D}) operations needed to transform $C$ to $B$. $E(B)$  can easily be computed by the edit distance function between $C$ and $B$. For example, \texttt{brangelina} is encoded as \texttt{CCCDDDCCCCCCC}. Since $E(B)$ has the same length of $C$, we can now model this as a \emph{fixed length sequence labeling} problem rather than a variable length sequence to sequence learning problem completely obviating the need for the \emph{``encoder-decoder''} architecture for this large class of blends.   
\end{enumerate}

\textbf{Equivalent Problem Definition} Given a string $C=C_{1}\#C_{2}$, consisting of components $C_{1}$ and $C_{2}$, learn $E(B)$, a labeling of each character in $C$ from the label set $\{\textbf{C}, \textbf{D}\}$.

\subsubsection{Proposed model: \textsc{CopyCat}}
\begin{figure}[t!]
	\includegraphics[scale=0.5, width=\columnwidth]{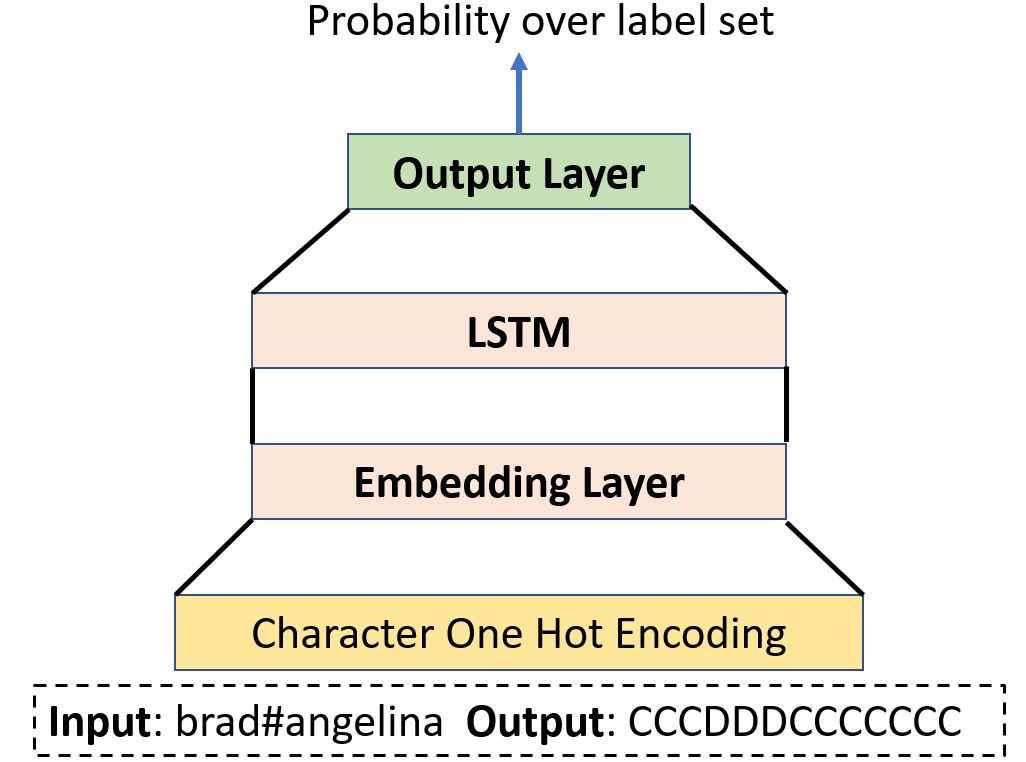}
	\caption{Our simple LSTM sequence labeling model for blends. Our model is considerably simpler both in complexity of the parameter space and in terms of implementation compared to the encoder-decoder model proposed by \cite{gangal2017charmanteau} which modeled the problem as a machine translation problem.}
	\label{fig:lstm}
\end{figure}
\paragraph{Model Architecture} Inline with work on neural sequence labeling \cite{wang2015part,plank2016multilingual}, our model uses a single layer long short term memory (LSTM) \cite{graves2012supervised} and an embedding layer as depicted in Figure \ref{fig:lstm}. Since the size of the training set is relatively small, we reduce the number of learn-able parameters by using pre-trained character embeddings frozen during training. In particular, we use $50$ dimensional character embeddings from \cite{gangal2017charmanteau} obtained by training an LSTM language model on an English Dictionary. The implementation of the LSTM layer is described by \cite{graves2012supervised,wang2015part} and therefore omitted here. The output layer is a soft-max over the label set \{\textbf{C, D}\}. 

\paragraph{Candidate Generation}
Our model outputs probability scores over the label set for each element in the sequence. As in previous works \cite{gangal2017charmanteau}, we use the output to generate an ordered candidate set $\mathcal{T}$. To construct $\mathcal{T}$, we use a simple top-K decoder which selects the $k$ most probable label sequences. Finding the $k$ most probable tag sequences from the soft-max outputs can be cast as finding the top $k$ shortest simple paths in a directed acyclic graph which can be efficiently solved using \cite{yen1971finding,eppstein1998finding}. Note that the greedy decoder which just picks the most likely label at each position is a special case with $k=1$. While the number of candidates generated by \cite{gangal2017charmanteau} depends on the size of $C$, our model generates a constant number of candidates ($k=5$) regardless of the input $C$.

\begin{table*}[htb]
\centering
\small
	\begin{tabularx}{\linewidth}{l|X|X|X|X}
	\textbf{Input} & \textbf{LSTM} & \textbf{LSTM + LM} & \textbf{LSTM + LM + LEN} & \textbf{Gold standard} \\
	\hline
	\texttt{brad\#angelina} & \texttt{brngelina} & \texttt{brangelina} & \texttt{brangelina} & \texttt{brangelina} \\
	\texttt{animated\#matrix} & \texttt{animtrix} & \texttt{animatrix} & \texttt{animatrix} & \texttt{animatrix} \\
	\texttt{merkel\#sarkozy} & \texttt{merkrkozy} & \texttt{merkkozy} & \texttt{merkkozy} & \texttt{merkozy} \\
	\texttt{dramatic\#drastic}  & \texttt{draastic} & \texttt{drastic} & \texttt{dramastic} & \texttt{dramastic} \\
	\texttt{kentucky\#indiana} & \texttt{keiana} & \texttt{keana} & \texttt{keniana} & \texttt{kentuckiana} \\
     \texttt{employability\#agility} & \texttt{emplglity} & \texttt{emplgility} & \texttt{employgility} & \texttt{employagility} \\
%	\texttt{education\#entertainment} & \texttt{educntertainment} & \texttt{eduntertainment} & \texttt{edutertainment} & \texttt{edutainment} \\
	\hline
	\end{tabularx}
	\caption{Set of examples demonstrating the effect of incorporating the language model and length model when ranking candidates revealing insights into blend formation. Incorporating the language model generally improves the fluency of the blend while the length model helps generate blends with lengths close to the target.}
	\label{tab:ranking_models}
\end{table*}

\begin{figure}[htb!]
	\includegraphics[scale=0.5, width=\columnwidth]{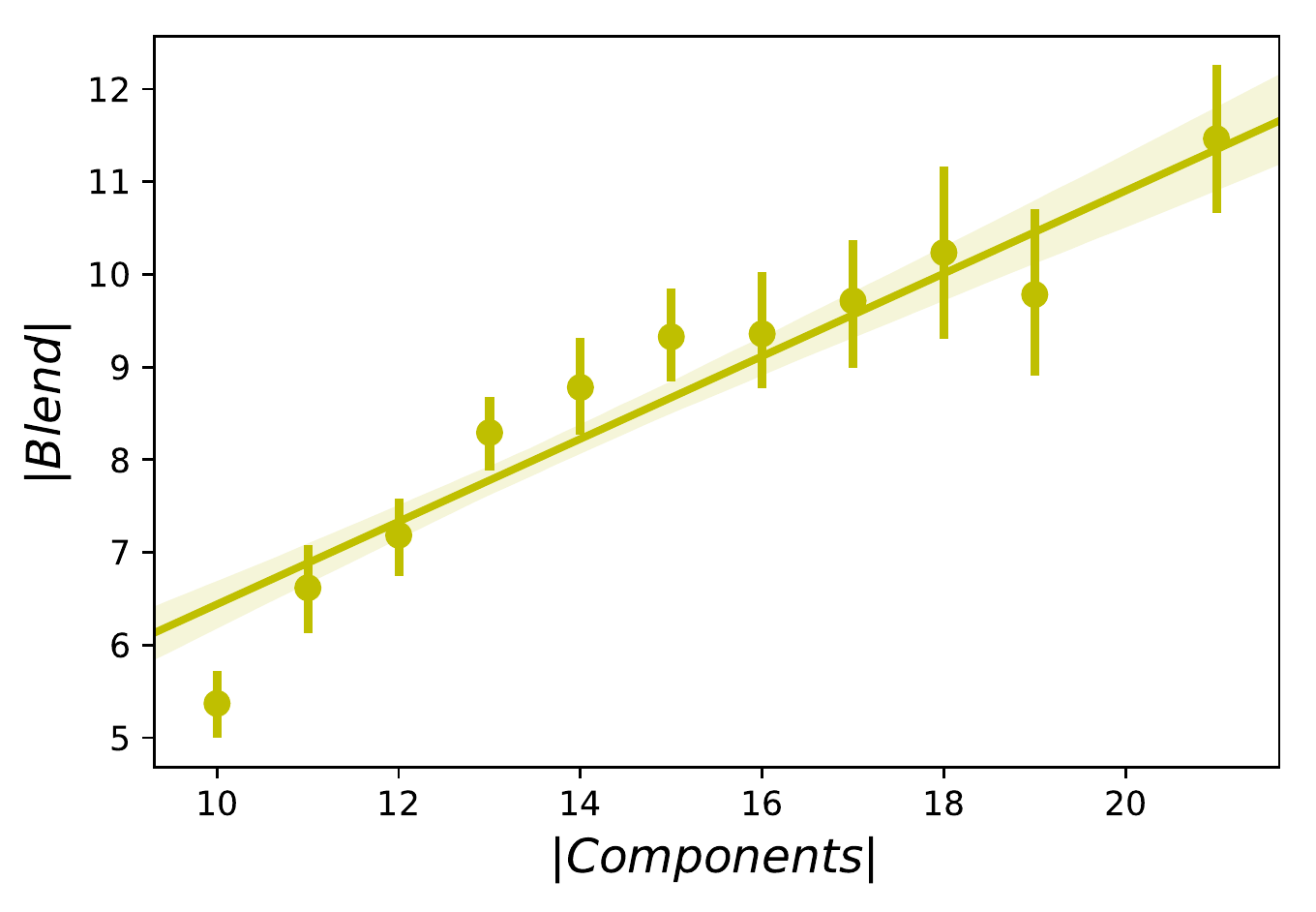}
	\caption{Best fit between blend length and the length of the input string, implying that one can infer the length of the blend from the length of the components using a regression model.}
	\label{fig:blends_reg_plot}
\end{figure}

\paragraph{Candidate Ranking and Selection} While the list of candidates in $\mathcal{T}$, can be used to make a prediction we note that re-ranking these candidates can result in better performance and thus consider multiple ranking strategies:
\begin{enumerate}
    \item \textbf{LSTM.} We consider only the scores as obtained by the LSTM with no re-ranking.
    \item \textbf{LSTM + LM.} We augment the score of each candidate to include both the score of the LSTM as well as its score according to a character level language model where the language model is trained on a large amount of unsupervised text \footnote{We use words from the CMU pronouncing dictionary.}.
    \item \textbf{LSTM + LM + LEN.} Figure \ref{fig:blends_reg_plot} shows a least squares fit to the length of the blends versus the length of its component, suggesting a strong correlation between these two variables. We capture this notion through a probabilistic model. Specifically, we model $\mathbf{Pr}(Blend_{len}|Component_{len})$ by fitting a Bayesian Ridge Regression model to the training data and score each candidate on this model as well. Finally, we combine the scores obtained for the LSTM, the language model and the length model uniformly to yield the final score for each candidate. 
\end{enumerate}
In each of the above cases, we pick the topmost candidate as our prediction.

Table \ref{tab:ranking_models} illustrates the effects of each of these ranking strategies on an exemplary set of strings. While ranking using the raw scores of the LSTM yields blends which are close, observe that the LSTM alone does not capture ease of pronunciation effectively. For example, the top ranking candidate \texttt{brngelina} is relatively unlikely under both a character language model and a phoneme language model. Incorporating scores from the language model results in much more natural blends like \texttt{brangelina}. To observe the effect of the length model, note that for \texttt{kentucky\#indiana} the blends obtained by LSTM and LSTM + LM are too short (\texttt{keiana} and \texttt{keana}). Incorporating the length model boosts scores for candidates closer to the target length yielding \texttt{keniana} which is closer to the target \texttt{kentuckiana}.  

\subsubsection{Evaluation} We compare our model to previous methods namely \cite{deri2015make} and \cite{gangal2017charmanteau}. Inline with \cite{gangal2017charmanteau}, we evaluate our model on $\mathcal{D}_{knight}$ and $\mathcal{D}_{blind}$ (consisting of $1078$ instances)\footnote{We discard relatively rare instances with insertions.}. For evaluating on $\mathcal{D}_{knight}$, we use $10$-fold cross-validation. For evaluating on $\mathcal{D}_{blind}$ we train our model on $\mathcal{D}_{knight}$ and report the mean score on the test-set obtained using $10$ random splits of the training data.  As in previous work, our metric is the edit distance between the predicted blend and the true blend. 

\begin{comment}
\begin{figure}[]
	\begin{subfigure}{0.48\linewidth}
		\includegraphics[width=\columnwidth]{figs/blends_prefix_length.pdf}
		\caption{Length of prefix}
		\label{fig:blends_prefix_length}
	\end{subfigure}
	\begin{subfigure}{0.48\linewidth}
		\includegraphics[width=\columnwidth]{figs/blends_suffix_length.pdf}
		\caption{Length of suffix}
		\label{fig:blends_suffix_length}
	\end{subfigure}
	\caption{Performance of our models for generating reduplicatives on both datasets.}
	\label{fig:blends_insight}
\end{figure}
\end{comment}

\subsection{Clippings}
\paragraph{Problem Formulation} Given a word $w$, learn a model that can generate its clipping $c$. For example, given the word \texttt{administration} we would like the model to output the clipping \texttt{admin}. 
\paragraph{Proposed Models} 
\begin{figure}[tb!]
	\includegraphics[scale=0.5, width=\columnwidth]{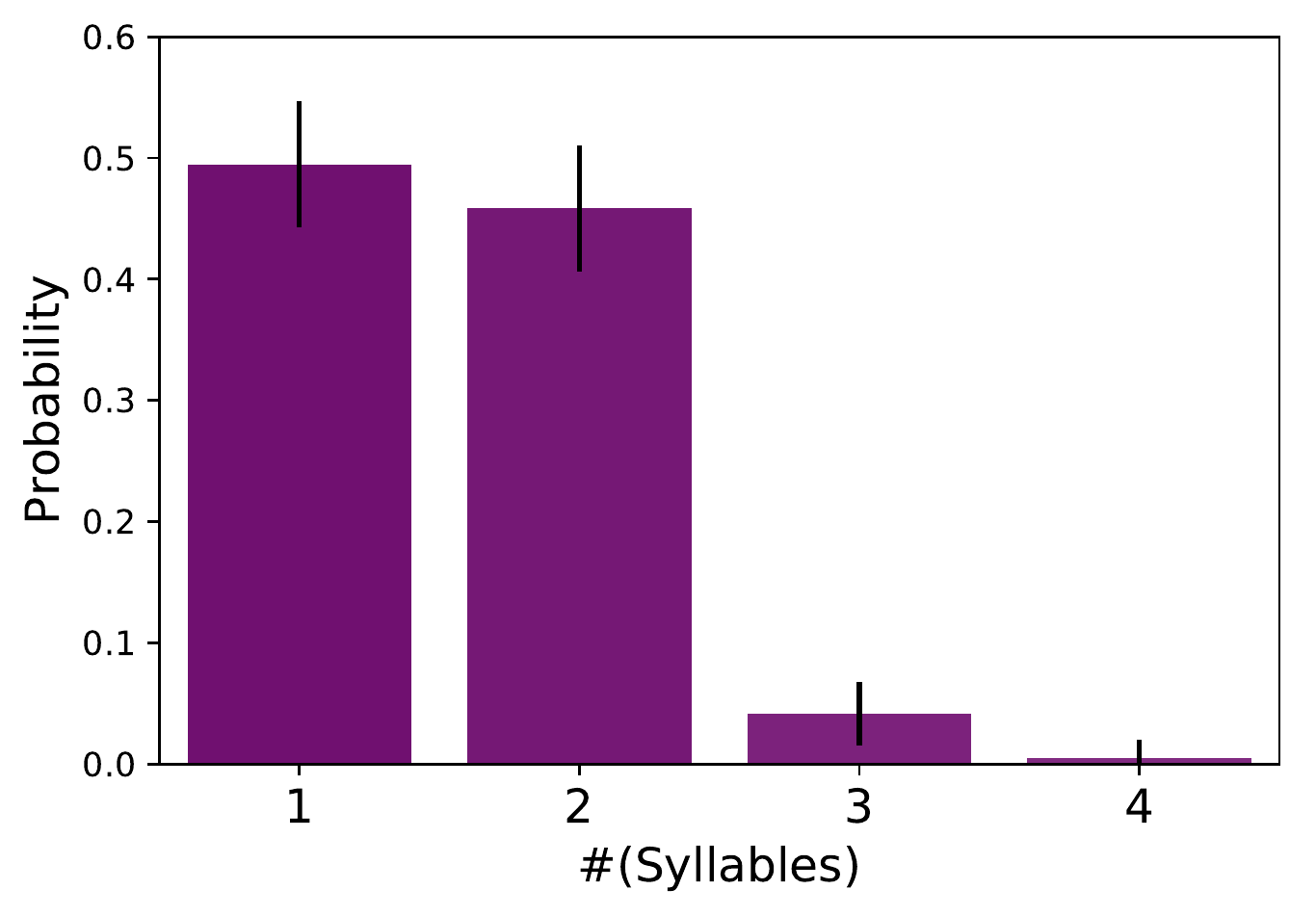}
	\caption{Distribution of the number of syllables in clippings. Most clippings have at-most 2 syllables but it is a challenge to infer whether a given word has a one or two syllable clipping.}
	\label{fig:clipping_num_syllables}
\end{figure}

We motivate our models by presenting two insights into linguistic properties of clippings first noted by \cite{mattiello2013extra}. First, most clippings are back clippings while fore clippings are relatively rare. Second, most clippings tend to have at most two syllables (see Figure \ref{fig:clipping_num_syllables}). The first insight guides our model in determining whether to retain a prefix or a suffix of the original word. The second insight guides our model in determining how much to retain (or clip). In particular, we can capture this in two ways: (a) Working in the phoneme space and (b) Learn a function to predict the length of the clipping (which encodes the number of syllables implicitly).

\subsubsection{\textsc{ClipPhone}}
\textsc{ClipPhone} operates by mapping the word $w$ to a sequence of phonemes, explicitly clipping in the phoneme space and mapping the phoneme sequence back to the grapheme space. In particular, the model can be described as follows: \textbf{(1)} Let $\theta$ and $\pi$ represent multinomial distributions over clipping types and the number of syllables respectively. \textbf{(2)} Represent the word $w$ as a sequence of phonemes  $\mathcal{P}$ and identify each syllable. \textbf{(3)} Draw a sample $l$ from $\pi$ to represent the number of syllables in the clipping and the type $t$ by drawing a sample from $\theta$. \textbf{(4)} If $t\in \{\textsc{Back, Fore}\}$, clip $\mathcal{P}$ to have exactly $l$ syllables by selecting the appropriate length prefix or suffix depending on clipping type $t$ to make exactly $l$ syllables represented by $\mathcal{P}_{clip}$. If $t$ is a \textsc{Compound}, clip each word recursively and concatenate the outputs. \textbf{(5)} Map $P_{clip}$ back to grapheme space to yield the clipping. \textbf{(6)} The parameters of $\theta$ and $\pi$, both multinomial distributions can be estimated from observed data via maximum likelihood estimators.
\paragraph{Phoneme-based representation} 
Given  $w$, we obtain its phoneme representation using the pre-trained state-of-art neural model \textsc{G2PSeq2Seq} \cite{yao2015sequence}\footnote{G2PSeq2Seq has a PER of $5.45\%$ on CMUDict.}. For example, the word \texttt{captain} is mapped to the following sequence of phonemes $\mathcal{P}$ given by \texttt{K AE P T AH N}. 
\paragraph{Clipping in the phoneme space} We identify the syllable boundaries in $\mathcal{P}$ by looking for vowel phonemes and clip $\mathcal{P}$ until it contains the desired number of syllables. For example, a one syllable clipping of \texttt{K AE P T AH N}  is \texttt{K AE P T} since \texttt{AH} is the beginning of the second syllable ($\mathcal{P}_{clip}=\texttt{K AE P T}$).
\paragraph{Mapping clipping back to graphemes}
Finally, given the clipped phoneme sequence $\mathcal{P}_{clip}$, we map it back to the grapheme space by learning a sequence to sequence model to map a phoneme sequence to its grapheme sequence. We follow the same model architecture as used in \textsc{G2PSeq2Seq} \cite{yao2015sequence} but just flip the input and output sequences, thus learning a model to map phonemes back to graphemes. Finally, we quantify the effectiveness of this model explicitly in our evaluation by establishing an upper-bound on the expected performance using this model.
\subsubsection{\textsc{ClipGraph}} As we will demonstrate empirically, \textsc{ClipPhone} poses the following challenges: Since a given phoneme sequence can map to multiple grapheme sequences, explicitly mapping to phoneme space and back introduces errors and loss of fidelity \footnote{\texttt{K AE P T} can be mapped to \texttt{capped} or \texttt{capt}.}. This is compounded by noting that whether $w$'s clipping should have one or two syllables is hard to predict (since both occur in almost equal proportions empirically, see Figure \ref{fig:clipping_num_syllables}). Furthermore, deciding whether the clipping should end in a vowel or a consonant and determining the length is yet another challenge. As an alternative, we propose \textsc{ClipGraph} where we seek to directly learn a model to predict the length of the clipping directly given $w$, obviating the need to work in phoneme space. \textsc{ClipGraph} works as follows: \textbf{(1)} Let $\theta$ represent multinomial distributions over clipping types and let $\pi$ be a model that predicts the length $l$ of the clipping ${w}_{clip}$ given $w$. \textbf{(2)} Draw a sample from $\theta$ to get the type of clipping $t$. \textbf{(3)} Retain the appropriate $l$-length prefix or suffix of $w$ based on the type $t$ handling compound words recursively. \textbf{(4)} We use Ridge regression to learn $\pi$ from the training data and estimate $\theta$ using its maximum likelihood estimator. 

\paragraph{Evaluation} To evaluate our models, we consider a naive baseline \textsc{Naive} which clips $w$ to one of its prefixes or suffixes randomly. For \textsc{ClipPhone}, we also consider one (\textsc{ClipPhone-1Syl}) and two (\textsc{ClipPhone-2Syl}) syllable clippings. Since \cite{jamet2009morphophonological} notes that determining whether a word has a one or two syllable clipping is extremely challenging, we also consider a model with an oracle (\textsc{ClipPhone(O)}) on the number of syllables in the clipping to quantify this. Finally,  we establish an upper bound (\textsc{G2P-Gold}) on the performance of any method using our learned P2G model. \textsc{G2P-Gold} maps the gold standard $G$ to phoneme space and maps the resulting phoneme sequence back to yield $\hat{G}$ as the predicted clipping. We use the edit distance between the predicted clipping and the gold standard as the evaluation metric.

\subsection{Reduplicatives}
\label{sec:method_redups}
\paragraph{Problem Formulation}
Given a word $v$, we seek to generate a word $w$ such that $v.w$ is a reduplicative. For example, given the word \texttt{flip}, we would like to generate \texttt{flop} or \texttt{flap} to yield the reduplicatives \texttt{flip-flop} or \texttt{flip-flap}.

\paragraph{Proposed Model}
We motivate our model from a linguistic standpoint by referring to observations by \cite{mattiello2013extra} that a large fraction of reduplicatives can be formed by either (a) Duplicating the word \textsc{Duplicate} (\texttt{boo-boo}) (b) Exchanging the initial vowel \textsc{VowelEx} (\texttt{bing-bong}) and (c) Exchanging the initial consonant \textsc{ConEx} (\texttt{teenie-weenie}). Other patterns include adding a consonant (\texttt{artsy-fartsy}) or adding \texttt{schm/shm} (\texttt{moodle-schmoodle}). In our work, we propose a generative model for the three dominant forms of reduplication mentioned above. Broadly, our model captures the notion that vowels and consonants display strong replacement preferences. For example, the vowel \texttt{i} is much more likely to be replaced with \texttt{a} than \texttt{u} (instances like \texttt{clip-clap, wishy-washy}). Similarly the consonant \texttt{t} is much more likely to be replaced by \texttt{w} (as in \texttt{teenie-weenie, tinky-winky}). We incorporate these insights into our generative model as follows: \textbf{(1)} Let $\theta$ be a distribution over the three different types of reduplicatives. Let $\phi_{v}, \psi_{c}$ be distributions over letters that replace vowel $v$ and consonant $c$ respectively. \textbf{(2)} Sample the type of reduplicative $t$ generated from $\theta$. \textbf{(3)} If $t$ is \textsc{Duplicate}, set $w$ to $v$ and return $w$ as the reduplicative component. \textbf{(4)} If $t$ is \textsc{VowelEx}, find the first vowel $x$ with non-zero replacement probability. Sample the replacement $z$ from $\phi_{x}$ and replace $x$ with $z$. If $t$ is \textsc{ConEx}, find the first consonant $c$ with non-zero replacement probability and sample the replacement $z$ from $\psi_{c}$. Replace $c$ with $z$. \textbf{(5)} Return the edited string as the second component of the reduplicative. \textbf{(6)} The parameters of multinomial distributions $\theta$, $\phi_{v}$, and $\psi_{c}$ are estimated via MLE estimates from the data.

\paragraph{Evaluation}
We consider two baseline models (a) \textbf{\textsc{Let}} Uniformly replace a letter with another letter in $v$ to return $w$. (b) \textbf{\textsc{Let(Cond)}} Uniformly replace a letter (vowel or consonant) with a letter from its class. Since reduplicatives are characterized by phonologically similar sounds, merely using edit distance as a metric for evaluation would be ineffective. For example, even the ill-rhyming \texttt{flip-flsp} has the same edit distance as the correct reduplicative \texttt{flip-flop}. Thus, we use a distance measure (MIR) defined over the phoneme space \cite{hixon2011phonemic} which effectively captures the similarity of two phonetic sequences by modeling the affinities between pairs of phonemes using a point access mutation matrix and is superior over metrics like PER (phoneme error rate).

%% file: experiments.tex
\section{Experiments}
Having described our models and evaluation metrics in the previous section, we proceed to evaluate our models empirically and describe our results. 
\paragraph{Blends} We train our model on $\mathcal{D}_{knight}$ and evaluate on the $\mathcal{D}_{blind}$ dataset as in \cite{gangal2017charmanteau} comparing against previous methods \cite{deri2015make,gangal2017charmanteau}. We set the number of hidden units to $50$ with a dropout probability of $0.5$\footnote{All hyper-parameters were chosen using a validation set.}. We use \textsc{Adam} \cite{kingma2014adam} optimizer with an initial learning rate of $0.001$ to train the model for $500$ epochs with early stopping over a validation set.

Tables \ref{tab:f1_gold_knight} and \ref{tab:f1_gold_blind} show the mean edit distance of our predictions from the target blend. First, the evaluation on the $\mathcal{D}_{knight}$ dataset compares the performance of our models against previous baselines. Note that just using our basic model \textsc{CopyCat -(LSTM + LM)} outperforms the baseline proposed by \cite{deri2015make} ($1.59$ vs $1.49$). Furthermore, observe that even our vanilla model (LSTM) significantly outperforms the equivalent ``FORWARD'' models by \cite{gangal2017charmanteau}\footnote{We obtain the numbers for these baselines from the latest released data and predictions at https://github.com/vgtomahawk/Charmanteau-CamReady.} that use greedy and beam-search decoding ($1.90$ and $2.37$ vs $1.75$).  Moreover, observe that our model even achieves almost equivalent performance to the ``FORWARD'' state-of-art model which uses exhaustive decoding ($1.37$ vs $1.33$). Furthermore, our model achieves competitive performance with the more complex models proposed by \textsc{Gangal-Backward} which use exhaustive decoding. We emphasize that we can achieve competitive performance using a simpler model without using exhaustive decoding.  Similar observations can also be made for the evaluation of the $\mathcal{D}_{blind}$ data set (our best model yields a score of $1.91$ vs $1.77$). Altogether these observations suggest that even simple models with effective modeling of linguistic structure can perform competitively and even outperform overly complex models (see Table \ref{tab:qualitative_blends} for a few example predictions).

\begin{table}[htb!]
\centering
	\begin{tabular}{l|l}
	\textbf{Model} & \textbf{Distance} \\
	\hline
	 \textsc{Knight} & 1.59  \\
	 \hline
	 \textsc{Gangal-Forward (Greedy)} & 1.90 \\
	 \textsc{Gangal-Forward (Beam)} & 2.37 \\
	 \textsc{Gangal-Forward (SOTA)}$^\dagger$ & 1.33 \\
	 \textsc{Gangal-Backward (SOTA)}$^\dagger$ & 1.12 \\
	 \hline 
	 \textsc{CopyCat-(LSTM)} & 1.75 \\
	 \textsc{CopyCat-(LSTM + LM)} &  1.49 \\
	 \textsc{CopyCat-(LSTM + LM + Len)} & \textbf{1.40} \\
	 \textsc{CopyCat-(LSTM + LM + Len)}$^\dagger$ & \textbf{1.37} \\
	\hline
	\end{tabular}
	\caption{$10$-fold cross validation performance of our blending model \textsc{CopyCat} in terms of edit distance (lower is better) on $\mathcal{D}_{knight}$ dataset. $\dagger$ indicates ensemble approach using sub-samples of training data consistent with previous work. Our simpler model yields competitive performance (especially compared to the state of art forward model) without the need for exhaustive decoding (which the state of art uses), uses a smaller learn-able parameter set while effectively using linguistic insights into the blending process.}
	\label{tab:f1_gold_knight}
\end{table}

\begin{table}[htb!]
\centering
	\begin{tabular}{l|l}
	\textbf{Model} & \textbf{Distance} \\
	\hline
	 \textsc{Knight} & 2.10  \\
	 \hline
	 \textsc{Gangal-Forward (SOTA)}$^\dagger$ & 1.76 \\
	 \textsc{Gangal-Backward (SOTA)}$^\dagger$ & 1.77 \\
	 \hline 
	 \textsc{CopyCat-(LSTM)} & 2.27 \\
	 \textsc{CopyCat-(LSTM + LM)} &  2.13 \\
	 \textsc{CopyCat-(LSTM + LM + Len)} & \textbf{1.98} \\
	 \textsc{CopyCat-(LSTM + LM + Len)}$^\dagger$ & \textbf{1.91} \\
	\hline
	\end{tabular}
	\caption{Performance of our blending model \textsc{CopyCat} in terms of edit distance (lower is better) on $\mathcal{D}_{blind}$ dataset. $\dagger$ indicates ensemble approach using sub-samples of training data consistent with previous work. Our simpler model yields competitive performance without the need for exhaustive decoding, uses a smaller learn-able parameter set while effectively using linguistic insights into the blending process. To ensure the comparison is fair, numbers for the baselines were obtained by filtering the released predictions for these models to the same set of words our models were evaluated on.}
	\label{tab:f1_gold_blind}
\end{table}

\begin{figure}[]
	\includegraphics[scale=0.5, width=\columnwidth]{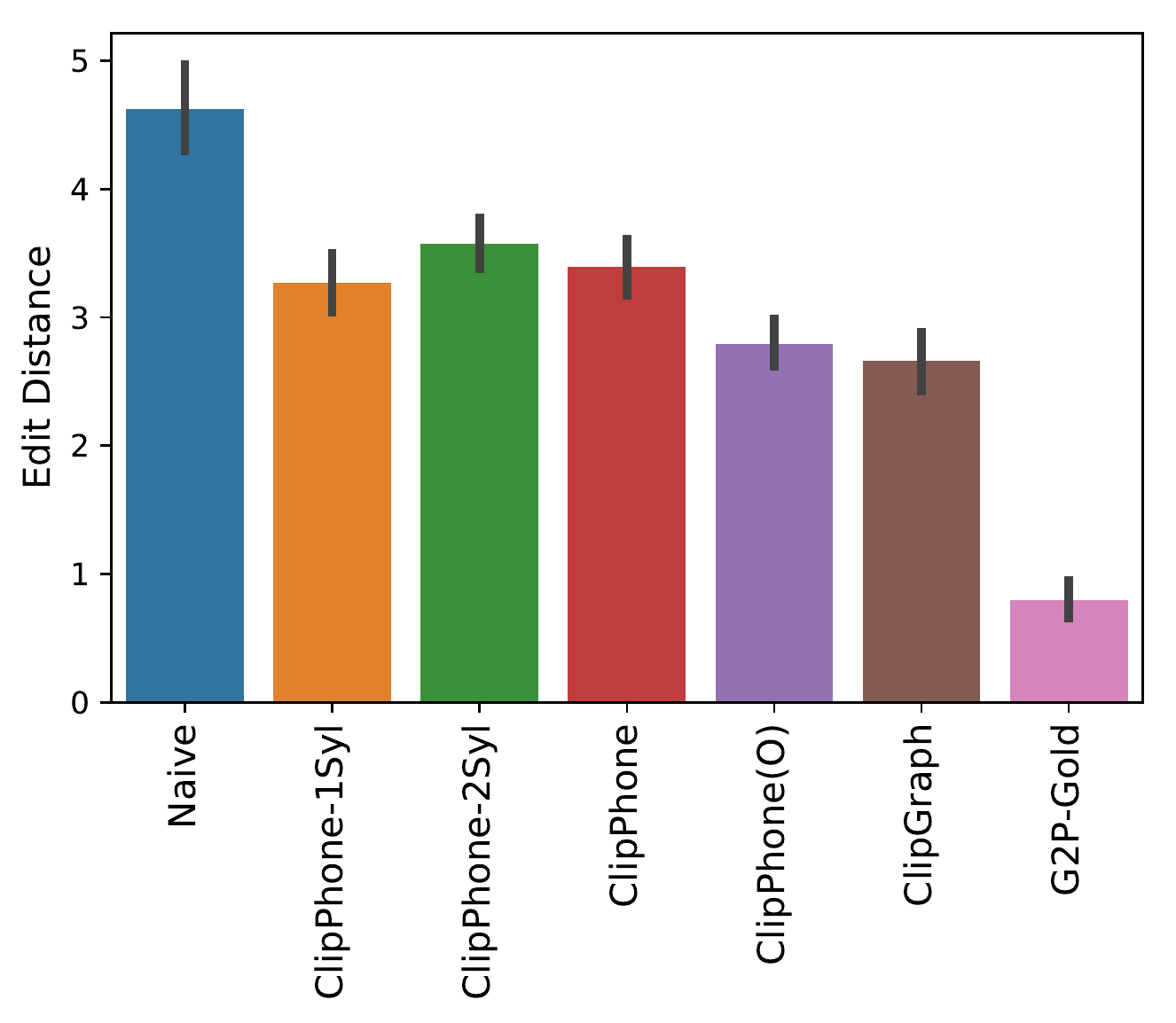}
	\caption{\textbf{Evaluation of clipping models} (lower is better). The \textsc{Naive} model is the performance lower bound. Both \textsc{ClipPhone} and \textsc{ClipGraph} substantially outperform the baseline. Estimating whether a given word has one or two syllable clipping is a major challenge hindering \textsc{ClipPhone} since both cases are equally likely from empirical estimation. Using an oracle on the number of syllables (\textsc{ClipPhone(O)}) improves performance. \textsc{ClipGraph} which operates purely in grapheme space performs as well as \textsc{ClipPhone(O)}. \textsc{G2PGold} denotes a upper bound when using our P2G model.}
	\label{fig:clipping_edit_distance}
\end{figure}

\paragraph{Clippings} We consider the dataset of clippings introduced by \cite{mattiello2013extra} and report the mean edit distance ($\mu$) on a set of $173$ clippings in Figure \ref{fig:clipping_edit_distance} (see Table \ref{tab:qualitative_clipping} as well). First, both \textsc{ClipPhone} ($\mu=3.39$) and \textsc{ClipGraph} ($\mu=2.65$) outperform the naive baseline ($\mu=4.6$). Second, based on the empirical distribution, it is almost impossible to predict whether a word has a one or two syllables ($0.49$ vs $0.46$ significant only at $\alpha>0.2$) clipping and incorrect guesses critically affect the downstream performance. In the absence of such information, just clipping to one syllable yields better performance. However, when this information is exactly known (\textsc{ClipPhone(O)}), we note an improvement as expected ($\mu=2.79$). Finally, \textsc{ClipGraph} shows a small but not significant advantage over methods working explicitly in the phoneme space which are prone to errors by imperfect conversion. 
Finally, the upper bound on the performance is substantially better than our models ($\mu=0.79$) suggesting scope for future improvements.

\begin{table}[htb!]
\small
\centering
	\begin{tabularx}{\linewidth}{l|l|l|l}
	 \textbf{Input} & \multicolumn{2}{c}{\textbf{Our models}} & \textbf{Gold} \\
	  & \textbf{\textsc{ClipPhone}} & \textbf{\textsc{ClipGraph}} & \textbf{Clipping} \\
	\hline
	\texttt{cocaine} & \texttt{coke} & \texttt{coca} & \texttt{coke} \\
	\texttt{juvenile} & \texttt{juve} & \texttt{juve} & \texttt{juvey} \\
	\texttt{amelia} & \texttt{umm} & \texttt{amel} & \texttt{mel} \\
	\texttt{alfred} & \texttt{fred} & \texttt{fred} & \texttt{fred} \\
	\texttt{kid video} & \texttt{kidvid} & \texttt{kivid} & \texttt{kidvid} \\
	\hline
	\end{tabularx}
	\caption{Exemplary predictions from clippings models. We can generate one/two syllable clippings, as well as compound clippings. Note the effect of incorrect P2G conversion for \texttt{amelia} as \texttt{umm} which is pronounced similar to \texttt{ame}. The current G2P model does not incorporate stress. It is possible that incorporating stress into the model can address this scenario.}
	\label{tab:qualitative_clipping}
\end{table}

\begin{table}[htb!]
\centering
\small
	\begin{tabularx}{\linewidth}{X|l|l|l}
	\textbf{Input} & \textbf{Prediction} & \textbf{Gold} & \textbf{Dist}\\
	\hline
	\texttt{jizz\#disney} & \texttt{jizzney} & \texttt{jizzney} & 0 \\
	\texttt{scum\#fuzz} & \texttt{scuzz} & \texttt{scuzz} & 0 \\
	\texttt{phone\#neck} & \texttt{phoneck} & \texttt{phoneck} & 0 \\
	\texttt{woman\#romance} & \texttt{womance} & \texttt{womance} & 0 \\
	\texttt{piss\#mishap} & \texttt{pisshap} & \texttt{pisshap} & 0 \\
	\texttt{pass\#asshole} & \texttt{passhole} & \texttt{passhole} & 0 \\
	\texttt{spike\#angel} & \texttt{spangel} & \texttt{spangel} & 0 \\
	\texttt{man\#amazon} & \texttt{mamazon} & \texttt{manazon} & 1 \\
	\texttt{awkward\#sauce} & \texttt{awkwauce} & \texttt{awksauce} & 1 \\
	\texttt{bed\#orgasm} & \texttt{bergasm} & \texttt{bedgasm} & 1 \\
	\hline
	\end{tabularx}
	\caption{Exemplary predictions from our simple blends model which suggests our model effectively captures blending phenomena by incorporating linguistic constraints.}
	\label{tab:qualitative_blends}
\end{table}

\begin{figure}[htb!]
		\includegraphics[width=\columnwidth]{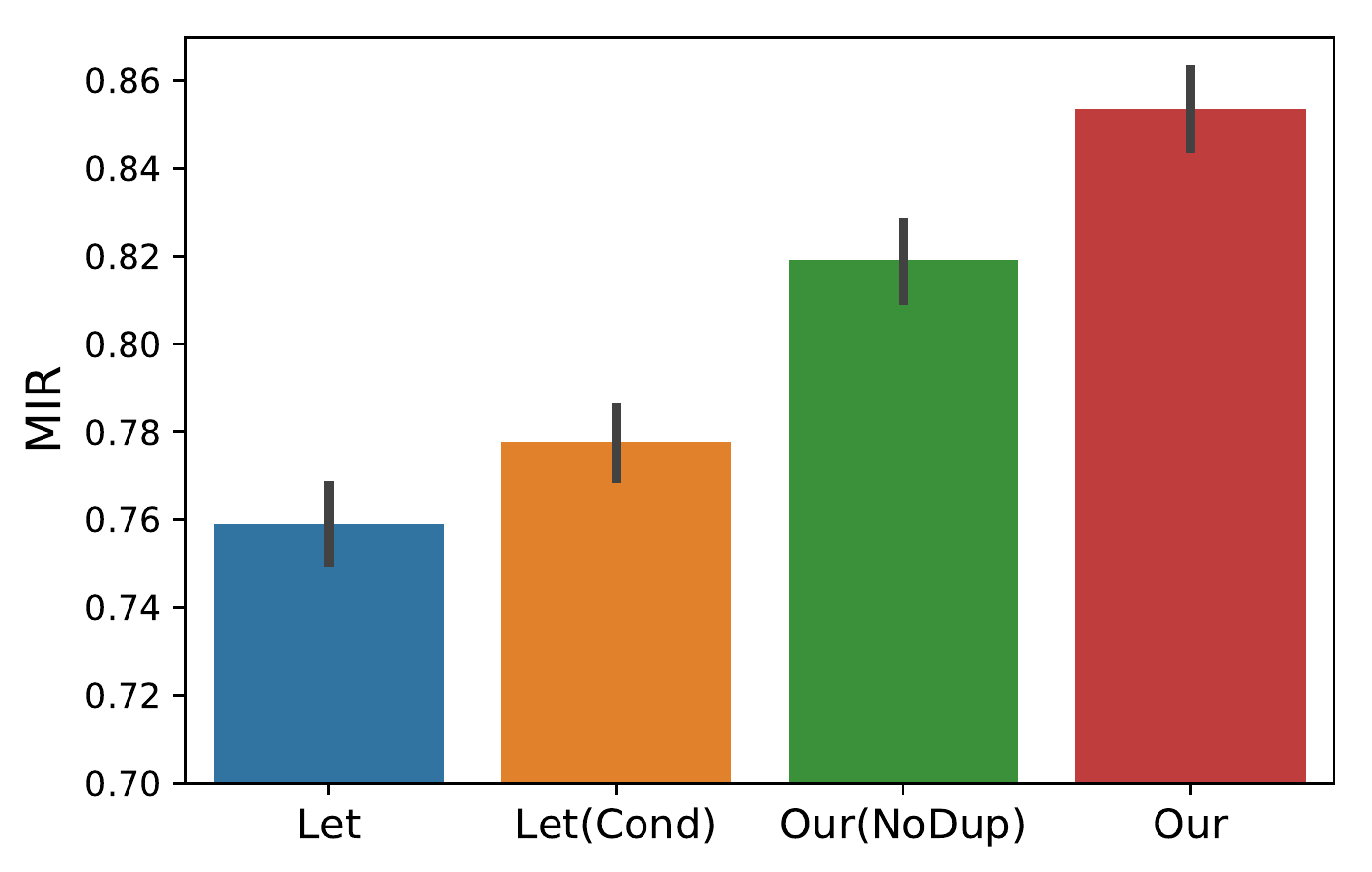}
	\caption{Performance of our reduplicative models based on the MIR metric (higher is better). Our models \textsc{Our(NoDup)} and \textsc{Our} consistently outperform baselines by at-least $8$ percentage points ($0.86$ vs. $0.78$).}
	\label{fig:redup_eval}
\end{figure}

\paragraph{Reduplicatives} We evaluate our model on a held-out test set of $50$ reduplicatives obtained using the manually compiled dataset by \cite{mattiello2013extra}. We evaluate two flavors of our model: (a) \textsc{Our(NoDup)} where we disallow generating duplicates (which are trivial to generate) and (b) \textsc{Our}, the full-fledged model where duplicate reduplicatives are allowed. We report the mean MIR for each model over $10$ independent runs in Figure \ref{fig:redup_eval}. Our model consistently outperforms the baselines (\textsc{Let}, and \textsc{Let(Cond)}) by at-least $8$ percentage points suggesting it adeptly captures patterns in reduplicative formation. Finally,  we examine the inferred probability distributions to gain insights into the linguistic phenomena in reduplicative formation few of which we outline: (a) The most common reduplicative types are  \textsc{Duplicate}, \textsc{VowelEx} followed by \textsc{ConEx}. (b) Vowel \texttt{i} is more  likely to be replaced by \texttt{a} and \texttt{o} and (c) Consonant \texttt{t} is much more likely to be replaced by \texttt{w} and \texttt{l} (like in \texttt{teenie-weenie}).  We leave a comprehensive analysis of these patterns to future work.

%% file: relatedwork.tex
\section{Related Work}
Blends, clippings and reduplicatives have been studied from a linguistic standpoint \cite{thun1963reduplicative,murata1990ab,merlini2008extra,hladky1998notes,hamans1997clippings,fandrych2008submorphemic,moehkardi2016patterns,ungerer2007word,beal1991toy,algeo1977blends,smith2014nehovah,shaw2014emergent,beliaeva2014unpacking,broad2016emergent,renner2013cross,gries2004isn,gries2004shouldn}. Most relevant is the work of \cite{mattiello2005pervasiveness,mattiello2008introduction,mattiello2013extra} who argues that slang is pervasive on the Internet, suggests its extra-grammatical nature and outlines some  phonological and morphological properties. Specifically, these phenomena are discussed followed by a qualitative analysis on a manually compiled dataset of $1580$ words from various sources. Recently \cite{deri2015make} and \cite{gangal2017charmanteau} study the problem of learning to blend and derive data-driven computational models for the task. \cite{deri2015make} propose a model based on finite state transducers. \cite{gangal2017charmanteau} outperform \cite{deri2015make} by modeling the problem of generating blends as a variable length sequence to sequence learning problem and propose a neural encoder-decoder based model. We differ from all of these works in several ways. In contrast to the blending model proposed by \citet{gangal2017charmanteau}, our model is simpler with lesser parameters, does not require using an encoder-decoder framework, or exhaustive decoding and yet yields competitive performance on a large class of blends.  We also propose the first computational generative data-driven models for clippings and reduplicatives while capturing phonetic similarity as well and evaluate our models quantitatively. 
 %Finally, we comprehensively evaluate of our models for all three phenomena, and publicly release data and models for further research.  

%% file: conclusion.tex
\section{Conclusion}
We proposed generative models for blends, clippings, and reduplicatives, three dominant word-formation phenomena in slang. Our models are distinguished by their simplicity, adept use of linguistic and structural constraints, easy to implement and yield state of the art performance. 

Our work suggests several directions for future research. First, our blending model can be extended to handle relatively rare insertions, incorporate the language model and the length model in a unified reinforcement learning framework optimizing a joint reward. Second, we do not investigate the complementary problem of de-blending. Moreover, we note that our evaluation of blends is based on an assumed gold standard. It would be useful to also characterize our blending model based on a human evaluation. Third, the gap between the performance of our clipping model and the upper bound (by the oracle) opens up the question of developing more nuanced models for clipping perhaps using deeper linguistic cues like stress patterns. Fourth, our models do not incorporate relatively rare types of reduplicative formation (like \emph{schm} reduplicatives) suggesting yet another direction for research. Yet another open question is whether a global model can effectively model all the above phenomena. Finally, in our work, we focus on only developing models for word formation in English slang. However, such word formation patterns are also evident in other languages \cite{vstekauer2012word} and it is an open question as to whether similar models generalize to other languages as well. 

Finally, our work potentially enables the development of several applications some of which include brand name generators, and rich conversational agents that are not only passive agents but can actively contribute to the evolution of language varieties. 

Altogether our work has implications for the broader fields of Internet Linguistics and natural language understanding especially in the context of slang formation.

%% file: acknowledgements.tex
\section*{Acknowledgements}
We thank members of the UCSB NLP Lab and the anonymous reviewers for their valuable comments and suggestions.